\documentclass[11pt,letterpaper]{article}
\usepackage{emnlp2018}
\usepackage{times}
\usepackage{latexsym}
\usepackage{tabu}
\usepackage[framemethod=tikz]{mdframed}
\usepackage{parsetree}
\usepackage{subcaption}
\usepackage{placeins}
\usepackage{appendix}
\usepackage{enumitem}

\aclfinalcopy

\newcommand{\mike}[1]{}
\renewcommand{\mike}[1]{{\color{red} [Mike: {#1}]}}
\newcommand{\mrinal}[1]{}
\renewcommand{\mrinal}[1]{{\color{red} [Mrinal: {#1}]}}
\newcommand{\sonal}[1]{}
\renewcommand{\sonal}[1]{{\color{red} [Sonal: {#1}]}}

\title{Semantic Parsing for Task Oriented Dialog \\ using Hierarchical Representations}

\author{Sonal Gupta \and Rushin Shah \and Mrinal Mohit \and Anuj Kumar \\ Facebook Conversational AI
        \AND
        Michael Lewis \\ Facebook AI Research
        \\ \{sonalgupta, rushinshah, mrinalmohit, anujk, mikelewis\}@fb.com}

\date{}

\begin{document}

\maketitle

\begin{abstract}
Task oriented dialog systems typically first parse user utterances to semantic frames comprised of intents and slots. Previous work on task oriented intent and slot-filling work has been restricted to one intent per query and one slot label per token, and thus cannot model complex compositional requests. Alternative semantic parsing systems have represented queries as logical forms, but these are challenging to annotate and parse. We propose a hierarchical annotation scheme for semantic parsing that allows the representation of compositional queries, and can be efficiently and accurately parsed by standard constituency parsing models. We release a dataset of 44k annotated queries \footnote{http://fb.me/semanticparsingdialog}, and show that parsing models outperform sequence-to-sequence approaches on this dataset.

\end{abstract}

\section{Introduction}
\begin{figure*}[ht!]
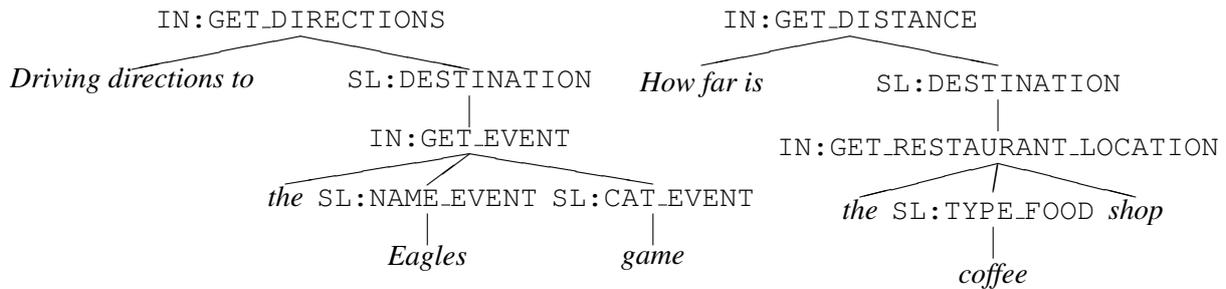
{}

    \begin{subfigure}[t]{0.3\textwidth}
        \begin{parsetree}
            \pthorgap{5pt}                         
            \ptvergap{12pt}                        
            \ptnodefont{\normalsize\rm}{7pt}{3pt}  
            \ptleaffont{\normalsize\it}{7pt}{3pt}  

            (.\texttt{IN:GET\_DIRECTIONS}. `Driving directions to' (.\texttt{SL:DESTINATION}. (.\texttt{IN:GET\_EVENT}. `the' (.\texttt{SL:NAME\_EVENT}. `Eagles') (.\texttt{SL:CAT\_EVENT}. `game') ) ) )
        \end{parsetree}
    \end{subfigure}%
    \hspace{0.2\textwidth}
    ~
    \begin{subfigure}[t]{0.3\textwidth}
        \begin{parsetree}
            \pthorgap{5pt}                         
            \ptvergap{12pt}                        
            \ptnodefont{\normalsize\rm}{9pt}{3pt}  
            \ptleaffont{\normalsize\it}{9pt}{3pt}  
            (.\texttt{IN:GET\_DISTANCE}. `How far is' (.\texttt{SL:DESTINATION}. (.\texttt{IN:GET\_RESTAURANT\_LOCATION}. `the' (.\texttt{SL:TYPE\_FOOD}. `coffee' ) `shop' ) ) )
        \end{parsetree}
    \end{subfigure}%

    \caption{Example TOP annotations of utterances. Intents are prefixed with \texttt{IN:} and slots with \texttt{SL:}.  In a traditional intent-slot system, the \texttt{SL:DESTINATION} could not have an intent nested inside it.}
    \label{fig:examples}
\end{figure*}

Intelligent personal assistants are now ubiquitous, but modeling the semantics of complex compositional natural language queries remains challenging.
Typical systems classify the \emph{intent} of a query (e.g. \texttt{GET\_DIRECTIONS}) and tag the necessary slots (e.g. \texttt{San Francisco})~\cite{mesnil2013,Liu2016AttentionBasedRN}.
It is difficult for such representations to adequately represent nested queries such as \emph{``Driving directions to the Eagles game"}, which is composed of \texttt{GET\_DIRECTIONS} and \texttt{GET\_EVENT} intents.
We explore a hierarchical representation for such queries, which dramatically improves the expressive power while remaining accurate and efficient to annotate and parse (see Figure \ref{fig:examples}).

We introduce a Task Oriented Parsing (TOP) representation for intent-slot based dialog systems.
This hierarchical representation is expressive enough to capture the semantics of complex nested queries, but is easier to annotate and parse than alternative representations such as logical forms or dependency graphs. We show empirically that our representation is expressive enough to model the vast majority of human-generated requests in two domains.

A key advantage of our representation is that it has a structure similar to standard constituency parses, allowing us to easily adapt algorithms developed for phrase structure parsing for inference. In particular, we use linear-time Recurrent Neural Network Grammars (RNNG)~\cite{dyer-rnng:16} and show that the inductive bias provided by this model significantly improves the accuracy compared to strong sequence-to-sequence (seq2seq) models based on CNNs, LSTMs and Transformers.

Our contributions in this paper are:
\setlist{nolistsep}
\begin{enumerate}
\item A hierarchical semantic representation for task oriented dialog systems that can model compositional and nested queries.
\item A publicly available dataset of 44k requests annotated with our representation. We show that our representation has very high coverage of these requests, and that inter-annotator agreement is high.
\item We show that the representation is learnable by standard algorithms. In particular, we show that the RNNG parsing model outperforms seq2seq baselines.
\end{enumerate}

\section{Representation}
Designing semantic annotation schemes requires trade-offs between how expressive the representation is on one hand, and how easily can it be annotated, parsed, and executed on the other. Most existing annotations for task oriented dialog systems have fallen on the extremes of non-recursive intent and slot tagging, such as in the ATIS dataset~\cite{mesnil2013,Liu2016AttentionBasedRN}, and full logical forms~\cite{zettlemoyer2012}.

We introduce a hierarchical representation, similar to a constituency syntax tree, with words as terminals. Non-terminals are either \emph{intents} or \emph{slots}, and the root node is an intent. We allow intents to be nested inside a slot, resulting in the ability to compose requests and call multiple APIs. Using this compositional tree representation, we can enable answering compositional queries over multiple domains.

We introduce the following constraints in our representation: 1. The top level node must be an intent, 2. An intent can have tokens and/or slots as children, 3. A slot can have either tokens as children or one intent as a child.

Executing queries such as those in Figure \ref{fig:examples} is straightforward because of the explicit tagging of the outer location slot: first we fetch `the Eagles game' event (or the relevant coffee shop), extract the location, and pass it as the destination slot to the navigation domain intent.

Compositional queries are frequent. In our dataset of crowd-sourced utterances, we found that 30\% could not be adequately represented with traditional intent-slot tagging.\footnote{In our dataset, 35\% of queries have depth $\textgreater 2$, which  means that the traditional intent-slot tagging systems would not have been  able to annotate or predict these annotations. In addition, to avoid  the depth-based statistic being influenced by our label set  specification, we manually performed analysis of 100 samples that showed  that 30\% of the queries required compositional representation.} This shows that more expressive representations are often necessary.

While our representation is capable of modeling many complex queries, some utterances are beyond its scope. For example, in \emph{Set an alarm at 8 am for Monday and Wednesday}, \emph{8 am} needs to be associated with both \emph{Monday} and \emph{Wednesday} which would require graph-structured representations. However, we found that just 0.3\% of our dataset would require a more expressive representation to model adequately.\footnote{Utterances  that could not be annotated with the label sets and corresponding  instructions were marked as `unsupported' (9.14\% of the dataset),  including the ones that needed more expressive representation than a tree.  Analysis of a random sample of 120 unsupported utterances shows  that only four instances cannot be supported because of the tree representation  constraint, estimating that there are only 0.3\% queries that would  require a more general (e.g graph-based) representation.} More expressive representations, such as dependency graphs or logical forms, would only bring marginal gains but would add significant challenges for annotation and learning.

Together, these results suggest that our tree-structured approach offers a useful compromise between traditional intent-slot tagging and logical forms, by providing very high coverage of queries while avoiding the complexities of annotating and learning more expressive representations.

In summary, our representation has the following attractive properties:
\begin{itemize}
    \item \textbf{Expressiveness} Compared to traditional intent-slot annotations, it can express complex hierarchical queries, improving coverage of queries by 30\%. We found that more general representations are required for only 0.3\% of queries.
    \item \textbf{Easy Annotation} Annotating our representation simply requires labeling spans of a sentence, which is much more straightforward than alternatives such as creating a logical form for the sentence, or an arbitrary dependency graph. This fact allows us to quickly create a large dataset.
    \item \textbf{Efficient and Accurate Parsing} Since our representation closely resembles syntactic trees, we can easily re-use models from the large literature on constituency parsing.
    \item \textbf{Execution} Our approach can be seen as a simple generalization of traditional dialog systems, meaning that existing infrastructure can easily be adapted to execute the intents.
\end{itemize}

\section{Dataset}
We asked crowdsourced workers to generate natural language sentences that they would ask a system that could assist in navigation and event queries. These requests were then labeled by two annotators. If these annotations weren't identical then we adjudicated with a third annotator. If all three annotators disagreed then we discarded the utterance and its annotations. 63.40\% of utterances were resolved with 2 annotations and 94.09\% were resolved after getting 3 annotations. We also  compared percentage of utterances that were resolved after 2 annotations for depth  $\leq 2$ (traditional slot filling) and for depth $> 2$ (compositional):  68.87\% vs 62.03\%, noting that the agreement rate is similar.

We collected a total of 44783 annotations with 25 intents and 36 slots, randomly split into 31279 training, 4462 validation and 9042 test utterances. The dataset has utterances that are focused on navigation, events, and navigation to events.

\begin{figure*}[htp!]
    \centering
    \begin{subfigure}{0.5\textwidth}
    \includegraphics[width=\textwidth]{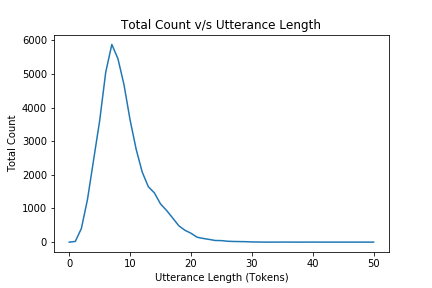}
    \end{subfigure}%
    \begin{subfigure}{0.5\textwidth}
    \includegraphics[width=\textwidth]{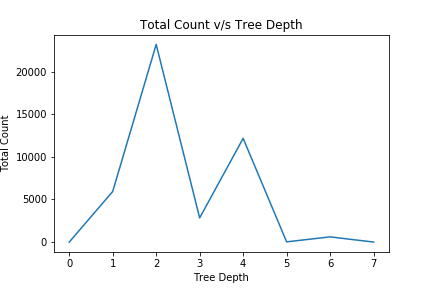}
    \end{subfigure}%
    \caption{Count statistics of the full dataset.}
    \label{fig:countstats}
\end{figure*}

The median (mean) depth of the trees is 2 (2.54), and the median (mean) length of the utterances is 8 (8.93) tokens. 35\% of trees have depth more than 2. The dataset has 4646 utterances that contain both navigation and event intents. Figure \ref{fig:countstats} shows the distribution of instances in the full dataset over the utterance length and tree depth.

\section{Models}
We experiment with two types of models: standard sequence-to-sequence learning models, and a model adapted from syntactic parsing, Recurrent Neural Network Grammars~\cite{dyer-rnng:16} (RNNG). RNNG is a top-down transition-based parser and was originally proposed for parsing syntax trees and language modeling.  We trained the RNNG parser discriminatively and not generatively to reduce training time of the model.  While sequence-to-sequence learning can model arbitrary sequence transduction, we hypothesize that parsing models like RNNG, which can only output well-formed trees, will give better inductive bias and flexibility for predicting compositional and cross-domains scenarios on the fly, particularly for domains with less training data available.

We briefly review the RNNG model -- The parse tree is constructed using a sequence of transitions, or `actions'. The transitions are defined as a set of SHIFT, REDUCE, and the generation of intent and slot labels. SHIFT action \textit{consumes} an input token (that is, adds the token as a child of the right most `open' sub-tree node) and REDUCE \textit{closes} a sub-tree. The third set of actions is generating non-terminals -- the slot and intent labels. Note that at each step, there are only a subset of valid actions. For examples, if all the input tokens have been added to the tree, the only valid action is REDUCE.



\section{Experiments and Results} \label{experiments}

\subsubsection*{Baselines}

We compared RNNG with implementations of sequence-to-sequence models using CNNs~\cite{gehring2017convs2s}, LSTMs~\cite{seq2seqlstm} and Transformer networks~\cite{Vaswani2017AttentionIA} in fairseq\footnote{https://github.com/pytorch/fairseq}.

\begin{table*}[ht!]
    \centering
    \resizebox{\textwidth}{!}{%
    \begin{tabular}{|l|c|c|c|c|c|c|c|c|}
        \hline
         Model & Exact match & $F_1$ & Precision & Recall & TL-$F_1$ & TL-Precision & TL-Recall & Tree Validity\\
         \hline
         \textbf{RNNG} & \textbf{78.51} & \textbf{90.23} & \textbf{90.62} & \textbf{89.84} & \textbf{84.27} & \textbf{84.64} & \textbf{83.91} & \textbf{100.00}\\
         seq2seq-CNN (LOTV) & 75.87 & 88.56 & 89.25 & 87.88 & 82.31 & 82.92 & 81.72 & 99.75\\
         seq2seq-LSTM (LOTV) & 75.31 & 87.69 & 88.35 & 87.03 & 81.15 & 81.72 & 80.58 & 99.94\\
         seq2seq-Transformer & 72.20 & 86.60 & 87.09 & 86.11 & 78.54 & 78.99 & 78.19 & 99.55\\
         \hline
    \end{tabular}}
    \caption{Performance (in percentage) of RNNG and seq2seq models based on LSTMs, CNNs, and Transformer networks. The CNN and LSTM models were trained with a Limited Output Token Vocabulary (LOTV) of just a single element.}
    \label{tab:comparison}
\end{table*}

\begin{figure*}[ht!]
    \centering
    \begin{subfigure}{0.5\textwidth}
    \includegraphics[width=\textwidth]{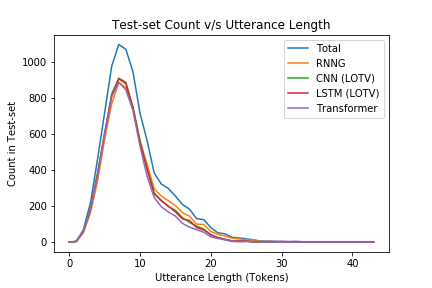}
    \end{subfigure}%
    \begin{subfigure}{0.5\textwidth}
    \includegraphics[width=\textwidth]{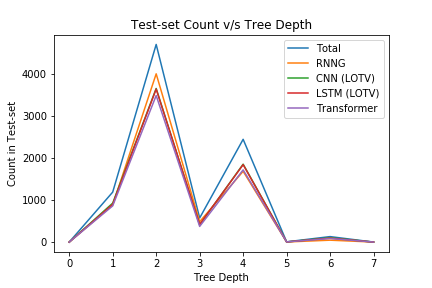}
    \end{subfigure}%
    \caption{Exact match statistics of the different models on the testset}
    \label{fig:fastats}
\end{figure*}

\subsubsection*{Metrics} We used three metrics to evaluate the systems. \textbf{Exact match} accuracy is defined as the number of utterances whose full trees are correctly predicted. The second metric is a commonly used scoring method for syntactic parsing -- \textbf{labeled bracketing $F_1$} scores~\cite{labeledbracketing} (called $F_1$ henceforth). We used the pre-terminals as well in the calculations. One downside of this metric is that it only considers the token spans for a given non-terminal but not the internal structure. Tree sub-structures are rather important for task completion. Thus, we introduce a third metric, \textbf{Tree-Labeled} (TL) $F_1$, which compares the sub-tree structure for a non-terminal, instead of just the token span. Exact match accuracy is the strictest metric and $F_1$ is the least strict metric. \textbf{Tree Validity} is the percentage of predictions which formed valid trees (via bracket matching).

\subsubsection*{Preprocessing and Hyperparameters}
We used the same preprocessing of unknown words as used in~\cite{dyer-rnng:16} and mapped numbers to a constant. We used pre-trained GloVe embeddings~\cite{pennington2014glove} and tuned hyperparameters on the validation set.
\begin{itemize}
    \item \textbf{RNNG} - We used 2-layer 164-unit LSTM with a bidirectional LSTM compositional function, trained with a learning rate of 0.0004, weight decay of 0.00004, dropout of 0.34 with Adam optimizer for 1 epoch with 16 workers using Hogwild updates.
    \item \textbf{seq2seq CNN} - We used 3x3 convolutions (9 layers of 512 units, 4 units of 1024 units), followed by 1x1 convolutions (2 layers of 2048 units) with attention, trained with initial learning rate of 0.9, dropout of 0.2, gradient clipping of 0.1 with NAG optimizer for 30 epochs, inferred with beam size 5.
    \item \textbf{seq2seq Transformer} - We used 3 layers of 4 FFN attention heads of embedding dimension 512, trained with initial learning rate of 0.01, dropout of 0.2, gradient clipping of 5 with Adam optimizer for 50 epochs, inferred with beam size 5.
    \item \textbf{seq2seq LSTM} - We used 1-layer 256 unit LSTMs with attention, trained with initial learning rate of 0.7, dropout of 0.2, gradient clipping of 0.1 with NAG optimizer for 40 epochs, inferred with beam size 5.
\end{itemize}

\subsubsection*{Results}
The experimental results in Table~\ref{tab:comparison} and Figure~\ref{fig:fastats} show that existing approaches for syntactic parsing, such as RNNG, perform well for this task, achieving perfect outputs on over 75\% of queries. RNNG performs better than sequence-to-sequence models, especially in predicting exact trees, which is important for task completion. We present results for the CNN and LSTM models with an output terminal token vocabulary of just a single element (LOTV), which performed better than the regular token vocabulary (exact match accuracy of 75.63\% and 68.39\%, respectively). We believe LOTV makes the models focus on learning to predict the tree structure rather than to reproduce the input tokens. But we observed that this vocabulary reduction resulted in significantly poorer performance for the Transformer model.  

Below we discuss how varying the beam size during inference affects accuracy. The seq2seq results in Table~\ref{tab:comparison} are accuracy for the top prediction when a beam of size 5 was run and the RNNG results are for greedy inference. For RNNG, the accuracy of top prediction did not change much when a beam of size 5 was run. We also measured how often the correct tree annotation was in the top $k$ predictions for the models when a beam of size $k$ was run during inference, called as \emph{Top-k}. For RNNG, \emph{Top-3} was 90.21 and \emph{Top-5} was 92.48, compared to 78.51 for \emph{k=}$1$. For seq2seq-CNN, the \emph{Top-3} score was 88.08 and \emph{Top-5} score was 90.21. For seq2seq-LSTM, \emph{Top-3} was 86.55 and \emph{Top-5} was 88.76. We note that \emph{Top-5} is substantially higher than the accuracy of the top prediction. These top-\emph{k} predictions could be used by a hypothesis ranker downstream, which can take into account agent capabilities.

We also experimented with more minimal representations of the RNNG model~\cite{rnngablation}. Removing the actions LSTM dropped Exact match score slightly to 78.08. Separately, removing the stack LSTM dropped it to 75.31. Removing the buffer LSTM caused a unusable decrease to 13.78.

While sequence-to-sequence models have shown strong parsing performance when trained on very large amounts of data~\cite{vinyalsgrammar}; in our setting the inductive bias provided by the RNNG model is crucial to achieving high performance. The model has several useful biases, such as guaranteeing a well-formed output tree, and shortening the dependencies between intents and their slots.

A further advantage of RNNG is that inference has linear time complexity, whereas seq2seq models are quadratic because attention is recomputed at every time step.


\section{Related Work}
Many annotation schemes have previously been proposed for representing the semantics of natural language. We briefly compare our method with these.

Most work on task oriented dialog systems has focused on identifying a single user intent and then filling the relevant slots -- for example, the representations used on the ATIS dataset \cite{mesnil2013, Liu2016AttentionBasedRN, 7953243} and in the Dialog State Tracking Challenge \cite{williams2016}. We showed that hierarchical representations with nested intents can improve coverage of requests in our domains.

The semantic parsing literature has focused on representing language with logical forms \cite{liangsemparsing,zettlemoyer2012,kwiatkowski2010}. Logical forms are more expressive than our representation, as they are less tightly coupled to the input query, and can be executed directly. However, logical forms are difficult to annotate and no large-scale datasets are available.

While we used a tree-structured representation, others have used arbitrary graphs, such as Abstract Meaning Representation \cite{banarescu2013abstract} and Alexa Meaning Representation \cite{alexamrl}. These approaches can represent complex constructions that are beyond the scope of our approach, but with significantly challenging parsing \cite{artzi2015}. We showed that such cases are very rare in our data.

\section{Conclusions}
Drawing on ideas from slot-filling and semantic parsing, we introduce a hierarchical generalization of traditional intents and slots that allows the representation of complex nested queries, leading to 30\% higher coverage of user requests. We show that the representation can be annotated with high agreement.
We are releasing a large dataset of annotated utterances at \url{http://fb.me/semanticparsingdialog}.
The representation allows the use of existing constituency parsing algorithms, resulting in higher accuracy than sequence-to-sequence models.

\appendix





\bibliographystyle{acl_natbib_nourl}

\end{document}